\documentclass[sigconf]{acmart}

\usepackage{subcaption}
\usepackage{color}
\usepackage{enumitem}
\usepackage{makecell}
\usepackage{algorithm}
\usepackage{listings}
\usepackage[table]{xcolor}
\setlist[itemize]{leftmargin=*}
\usepackage{multirow}
\usepackage{booktabs}
\definecolor{mygray}{RGB}{230,230,230}
\definecolor{mygreen}{RGB}{230,245,238}
\definecolor{myred}{RGB}{255,230,230}
\definecolor{textred}{RGB}{255,25,25}
\definecolor{textgreen}{RGB}{25,163,101}

\AtBeginDocument{%
  }

\copyrightyear{2026}
\acmYear{2026}
\setcopyright{cc}
\setcctype{by-nc-nd}
\acmConference[KDD '26]{Proceedings of the 32nd ACM SIGKDD Conference on Knowledge Discovery and Data Mining V.2}{August 09--13, 2026}{Jeju Island, Republic of Korea}
\acmBooktitle{Proceedings of the 32nd ACM SIGKDD Conference on Knowledge Discovery and Data Mining V.2 (KDD '26), August 09--13, 2026, Jeju Island, Republic of Korea}
\acmDOI{10.1145/3770855.3818845}
\acmISBN{979-8-4007-2259-2/2026/08}

\begin{document}

\title{Enhancing Protein Representation Learning via  \\ Manifold Restore Mixing}

\author{Yizhou Dang}
\affiliation{%
  \institution{Software College, Northeastern University}
  \city{Shenyang}
  \country{China}
}
\email{dangyz@mails.neu.edu.cn}

\author{Chuang Zhao}
\affiliation{%
  \institution{Tianjin University}
  \city{Tianjin}
  \country{China}
}
\email{zhaochuang@tju.edu.cn}

\author{Lianbo Ma}
\affiliation{%
  \institution{Software College, Northeastern University}
  \city{Shenyang}
  \country{China}
}
\email{malb@swc.neu.edu.cn}

\author{Guibing Guo}
\authornotemark[1]
\affiliation{%
  \institution{Software College, Northeastern University}
  \city{Shenyang}
  \country{China}
}
\email{guogb@swc.neu.edu.cn}

\author{Xingwei Wang}
\affiliation{%
  \institution{School of Computer Science and Engineering, Northeastern University}
  \city{Shenyang}
  \country{China}
}
\email{wangxw@mail.neu.edu.cn}

\author{Zhu Sun}
\authornote{Corresponding authors.}
\affiliation{%
  \institution{Information Systems Technology and Design, Singapore University of Technology and Design}
  \city{Singapore}
  \country{Singapore}
}
\email{sunzhuntu@gmail.com}

\renewcommand{\shortauthors}{Yizhou Dang, et al.}

\begin{abstract}
Data augmentation (DA) has been proven to be an effective means for improving protein representation learning (PRL) by generating additional training samples. Although mainstream perturbation- and sampling-based augmentation methods can produce data containing sufficient variations, they carry the risk of disrupting the protein structure and function. Some crafted protein homology modeling tools can generate conformations, but reduce structural diversity. 
The above dilemmas lead us to a question: Can we restore the disrupted structure caused by DA operations, providing data with both the original structure and diverse variations? In this work, we first analyze and empirically reveal the structure defect and performance degradation issues of existing DA methods. Based on the findings, we propose a simple yet effective DA method, \textbf{M}anifold \textbf{R}estore \textbf{M}ixing (MRM), for protein representation learning. Specifically, inspired by manifold mixup, we mix the hidden representations of original and augmented protein data to generate new samples that restore structural information lost in DA while introducing diverse variations. Furthermore, we develop a sample difficulty scheduler that adjusts the beta distribution in mixup to provide models with progressively challenging mixed samples during training, which improves the final performance. Comprehensive experiments on various PRL backbones and downstream tasks demonstrate the effectiveness and generalization of our method. The complete code and weights will be released upon acceptance. We provide a implementation at \url{https://github.com/KingGugu/MRM}. 
\end{abstract}





\maketitle

\section{Introduction}
\label{sec:intro}
Proteins are one of the essential biological entities for building cells and maintaining life activities, which consist of one or more 1D amino acid chains and perform various functions by folding into 3D conformations \cite{whitford2013proteins,watson2023novo}. Understanding and modeling proteins are significant for life sciences \cite{jumper2021highly,kim2025easy}. In recent years, deep learning has emerged as a powerful tool in protein understanding, which learns representations of protein structures and then uses them for a variety of tasks, such as protein design \cite{ovchinnikov2021structure,dauparas2022robust}, structure classification \cite{fan2023continuous,quan2024clustering}, protein folds quality assessment \cite{baldassarre2021graphqa}, and function prediction \cite{gligorijevic2021structure,zhou2022tasser}. Early approaches focused on modeling the 1D amino acid sequence of proteins \cite{vaswani2017attention,strodthoff2020udsmprot,brandes2022proteinbert}. Subsequently, some researchers explored learning protein representations with 3D geometric structures \cite{derevyanko2018deep,xia2021geometric}. More recently, methods that simultaneously model 1D and 3D structures have been proposed and demonstrated outstanding performance \cite{fan2023continuous,zhangprotein,wanglearning}.

\begin{figure}[!t]
  \centering
  \includegraphics[scale=0.815]{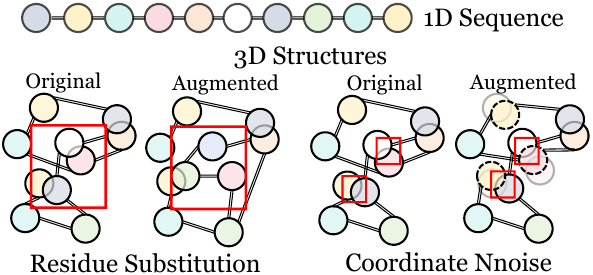}
  \vspace{-0.5em}
  \caption{Illustration of structure issues in the existing protein DA (highlighted in red boxes). The substitution alters the protein's folding pattern. The coordinate noise disrupts the original conformation. The spatial relationship between different residues has changed.}
  \label{fig:example}
  \vspace{-1em}
\end{figure}

Due to the challenge of experimental protein 3D structure determination, the 3D structural data available for model training is limited \cite{berman2000protein,zhangprotein}. To address this, many data augmentation methods have been proposed to produce data with diverse variations. The augmented data can be used to enhance model performance on downstream tasks \cite{sun2024enhancing,kim2025mixingdta} or for self-supervised pretraining \cite{zhangprotein,hermosilla2022contrastive}. For example, replacing amino acid residues \cite{sun2024enhancing}, introducing noise into the atomic coordinates \cite{fan2023continuous}, or sampling protein sub-structures \cite{hermosilla2022contrastive,hu2024deep}, etc. Despite the effectiveness, the above perturbation- and sampling-based augmentation operations may disrupt or lose the structure of the core functional domain in proteins, which is essential for model learning and prediction \cite{ovchinnikov2021structure,gligorijevic2021structure}. This issue ultimately prevents the model from effectively learning protein structures or establishing structure-function relationships, thereby compromising the performance of downstream tasks. For example, as illustrated in Figure \ref{fig:example}, if the substituted amino acid interacts differently with its neighbors, it may alter the protein's folding pattern, rendering it unable to perform its function. Coordinate perturbations may disrupt the original conformation, thereby affecting biological activity. Additionally, although protein homology modeling tools \cite{webb2016comparative,waterhouse2018swiss} or diffusion models \cite{fu2024latent,lu2025all} can generate protein conformations, some studies have indicated that these methods result in new data with limited structural diversity \cite{wangenhancing}.

The aforementioned predicament leads us to a question: \textbf{Can we restore the disrupted structure caused by DA operations, providing data with both the original structure and diverse variations?} In this work, we first review and analyze the structural issues that common augmentation operations may introduce. Furthermore, we empirically demonstrate the performance degradation they cause. Based on the limitations and findings, we propose a simple yet effective DA method, \textbf{M}anifold \textbf{R}estore \textbf{M}ixing (MRM), for protein representation learning. Inspired by the manifold mixup \cite{verma2019manifold,li2021feature}, we leverage representation-level mixing operations to fuse information from both original and augmented protein data to repair structural damage incurred during data augmentation in the latent space. This generates data that preserves original structures while incorporating rich variations. The original representation and the final mixed representation are jointly integrated into the model training process. Furthermore, we developed a difficulty scheduler that gradually reduces the proportion of raw representations in final mixed samples during training by adjusting the beta distribution of mixup, providing the model with samples ranging from easy to difficult. Lastly, we introduce a two-stage training strategy to avoid the noise and convergence instability introduced by the hybrid representation at the beginning of training. 

The main contributions can be summarized follows:
\begin{itemize}
    \item We highlight the defects of existing protein DA methods in preserving essential structural information. Analytical and empirical studies further substantiate our findings.
    \item We propose Manifold Restore Mixing, a manifold augmentation method that restores structural information lost in DA while introducing diverse variations to the augmented data. We equip MRM with a sample difficulty scheduler with two-stage regularized training, further improving performance.
    \item We conduct comprehensive experiments on multiple PRL backbones and downstream tasks to demonstrate the effectiveness and generalization of our method. It improves the performance of various typical PRL methods and outperforms existing data augmentation techniques \& protein language models.
\end{itemize}

\begin{table*}[!t]
  \centering
  \caption{Summary of existing data augmentation methods and their impact on structural information.}
  \vspace{-1em}
  \renewcommand\arraystretch{0.9}
  \scalebox{0.84}{
    \begin{tabular}{lccc}
    \toprule
    \multicolumn{1}{c}{Operation} & Object & Correspondence Between 1D and 3D & 1D and 3D Structure \\
    \midrule
    Deletion/Crop/Cut/Subsequence \cite{sun2024enhancing,zhou2025protclip} & \multirow{4}[2]{*}{1D Sequence} & No amino acid corresponds to the 3D coordinates & \multirow{3}[1]{*}{1D Incomplete, 3D do not apply} \\
    Swap/Shuffle/Reverse \cite{sun2024enhancing} &       & The original order has been disrupted &  \\
    Insert/Expansion/Contraction \cite{sun2024enhancing} &       & No 3D coordinates correspond to amino acids &  \\
    Substitution/Mask \cite{rao2021msa,zhang2023systematic,zhangprotein,sun2024enhancing} & & Retained & 1D Incomplete, some residues have been altered \\
    \midrule
    Torsion Angle Perturbation \cite{lan2025contrastive} & \multirow{3}[2]{*}{3D Structure} & \multirow{3}[2]{*}{Retained} & \multirow{3}[2]{*}{1D Retained, 3D has been altered} \\
    Gaussian Coordinate Noise \cite{hu2024deep,hu2024learning,lan2025contrastive} &       &       &  \\
    Rotation/Translation \cite{nguyen2025advances} &       &       &  \\
    \midrule
    Substructure Sampling \cite{zhangprotein,hermosilla2022contrastive,kalifa2025fusionprot,zhang2023systematic} & \multirow{2}[2]{*}{1D \& 3D} & \multirow{2}[2]{*}{Retrained} & \multirow{2}[2]{*}{Incomplete, partial 1D \& 3D structure loss} \\
    Subspace Sampling \cite{zhangprotein,kalifa2025fusionprot} &       &       &  \\
    \bottomrule
    \end{tabular}}%
  \label{tab:existing_aug}%
\end{table*}%

\begin{table*}[!t]
  \centering
  \caption{Performance of two representative backbones with different protein data augmentation operations. Results higher than the baseline model are highlighted in \textcolor{textred}{red}, while results less than or equal to the base model are highlighted in \textcolor{textgreen}{green}.}
    \vspace{-1em}
    \renewcommand\arraystretch{0.9}
  \scalebox{0.975}{
    \begin{tabular}{c|l|cccc|c|ccc|c}
    \toprule
    \multicolumn{2}{c|}{\multirow{2}[2]{*}{Method}} & \multicolumn{4}{c|}{Fold Classification} & Enzyme & \multicolumn{3}{c|}{Gene Ontology} & Enzyme \\
    \multicolumn{2}{c|}{} & Fold  & Superfamily & Family & Average & Reaction & BP    & MF    & CC    & Commission \\
    \midrule
    \multirow{7}[4]{*}{GearNet} & Base  & 28.4  & 42.6  & 95.3  & 55.4  & 79.4  & 0.356  & 0.503  & 0.414  & 0.730  \\
\cmidrule{2-11}          & w/ Substitution & \cellcolor{mygreen} 24.9  & \cellcolor{mygreen} 41.1  & \cellcolor{mygreen} 94.2  & \cellcolor{mygreen} 53.4  & \cellcolor{mygreen} 77.6  & \cellcolor{mygreen} 0.342  & \cellcolor{mygreen} 0.491  & \cellcolor{mygreen} 0.392  & \cellcolor{mygreen} 0.719  \\
          & w/ Mask & \cellcolor{mygreen} 28.2  & \cellcolor{myred}43.3 & \cellcolor{mygreen} 95.1  & \cellcolor{mygreen} 55.5  & \cellcolor{mygreen} 78.5  & \cellcolor{mygreen} 0.347  & \cellcolor{mygreen} 0.499 & \cellcolor{mygreen} 0.405 & \cellcolor{myred} 0.737 \\
          & w/ Torsion Angle Perturbation & \cellcolor{mygreen} 25.3  & \cellcolor{mygreen} 42.3  & \cellcolor{myred} 95.5 & \cellcolor{mygreen} 54.4  & \cellcolor{mygreen} 79.2  & \cellcolor{myred} 0.359 & \cellcolor{mygreen} 0.496 & \cellcolor{mygreen} 0.409 & \cellcolor{myred} 0.735 \\
          & w/ Gaussian Coordinate Noise & \cellcolor{mygreen} 28.1 & \cellcolor{myred} 42.9 & \cellcolor{mygreen} 94.8  & \cellcolor{mygreen} 55.3  & \cellcolor{myred} 79.8 & \cellcolor{mygreen} 0.353  & \cellcolor{mygreen} 0.501 & \cellcolor{myred} 0.416 & \cellcolor{mygreen} 0.726 \\
          & w/ Substructure Sampling & \cellcolor{mygreen} 27.1  & \cellcolor{mygreen}  41.8  & \cellcolor{mygreen} 93.9  & \cellcolor{mygreen} 54.3  & \cellcolor{mygreen} 78.7  & \cellcolor{mygreen} 0.350  & \cellcolor{mygreen} 0.487  & \cellcolor{mygreen} 0.398 & \cellcolor{mygreen} 0.722 \\
          & w/ Subspace Sampling & \cellcolor{mygreen} 27.7  & \cellcolor{mygreen} 42.3  & \cellcolor{mygreen} 94.5  & \cellcolor{mygreen} 54.8  & \cellcolor{myred} 79.6 & \cellcolor{mygreen} 0.339  & \cellcolor{mygreen}  0.482 & \cellcolor{mygreen} 0.401 & \cellcolor{mygreen} 0.714 \\
    \midrule
    \multirow{6}[4]{*}{CDConv} & Base  & 56.7  & 77.7  & 99.6  & 78.0  & 88.5  & 0.453  & 0.654  & 0.479  & 0.843  \\
\cmidrule{2-11}          & w/ Substitution & \cellcolor{mygreen} 47.5  & \cellcolor{mygreen} 75.4  & \cellcolor{mygreen} 98.9  & \cellcolor{mygreen} 73.9  & \cellcolor{mygreen} 87.3  & \cellcolor{mygreen} 0.444 & \cellcolor{mygreen} 0.649 & \cellcolor{mygreen} 0.463 & \cellcolor{myred} 0.849 \\
          & w/ Mask & \cellcolor{mygreen} 55.9  & \cellcolor{myred} 78.1 & \cellcolor{mygreen} 99.4  & \cellcolor{mygreen} 77.8  & \cellcolor{mygreen} 88.2  & \cellcolor{myred} 0.455 & \cellcolor{mygreen} 0.651 & \cellcolor{mygreen} 0.456 & \cellcolor{mygreen} 0.822 \\
          & w/ Torsion Angle Perturbation & \cellcolor{mygreen} 47.9  & \cellcolor{myred} 78.5 & \cellcolor{mygreen} 99.5  & \cellcolor{mygreen} 75.3  & \cellcolor{mygreen} 86.9  & \cellcolor{mygreen} 0.438 & \cellcolor{mygreen} 0.647  & \cellcolor{myred} 0.482 & \cellcolor{mygreen} 0.827 \\
          & w/ Substructure Sampling & \cellcolor{mygreen} 50.2  & \cellcolor{mygreen} 77.0  & \cellcolor{mygreen} 99.6  & \cellcolor{mygreen} 75.6  & \cellcolor{mygreen} 87.8  & \cellcolor{mygreen} 0.441 & \cellcolor{mygreen} 0.639 & \cellcolor{mygreen} 0.458 & \cellcolor{myred} 0.847 \\
          & w/ Subspace Sampling & \cellcolor{mygreen} 55.4  & \cellcolor{mygreen} 76.9  & \cellcolor{mygreen} 99.6  & \cellcolor{mygreen} 77.3  & \cellcolor{myred} 88.6 & \cellcolor{mygreen} 0.436 & \cellcolor{mygreen} 0.642 & \cellcolor{mygreen} 0.467 & \cellcolor{mygreen} 0.819 \\
    \bottomrule
    \end{tabular}}%
  \label{tab:defect}%
  \vspace{-1em}
\end{table*}%

\section{Related Work}
\label{sec:related}

\noindent \textbf{Protein Representation Learning.} PRL has attracted much attention in the fields of bioinformatics \cite{lin2023evolutionary,kim2025easy}. Early works focused on sequence-based PRL since proteins are chains of amino acids \cite{rao2021msa}. They leveraged the long short-term memory \cite{rao2019evaluating}, one-dimensional convolutional neural network \cite{shanehsazzadeh2020transfer}, and Transformer \cite{rao2021msa,gong2024evolution} to capture the sequential pattern and semantic relationships between residues. Later, structure-based methods emerged due to their ability to capture spatial relationships \cite{derevyanko2018deep,wangenhancing}. Another reason is that the function of a protein is determined by its structure. For example, 3DCNN \cite{derevyanko2018deep} used 3D atomic densities of proteins as input, employing alternating convolutional layers to assess the quality of protein folds. More recently, some researchers proposed methods that simultaneously capture sequential (1D) and spatial relationships (3D), achieving state-of-the-art (SOTA) performance \cite{zhangprotein,fan2023continuous,lee2023pre,hu2024learning}. CDConv \cite{fan2023continuous}, utilized continuous and discrete approaches to model the irregular and regular structures in proteins. ProNet \cite{wanglearning} captured hierarchical relations among different levels (amino acid, backbone, and all-atom levels). In addition, pre-trained models based on large-scale data \cite{rives2021biological,lin2023evolutionary} and clustering methods \cite{quan2024clustering} have also demonstrated satisfactory performance. Since the 1D sequence and 3D structure of a protein provide different types of information for model learning and achieve SOTA performance, our research focuses on methods that combine both 1D and 3D structures.

\vspace{0.3em}

\noindent \textbf{Mixup for Data Augmentation.}
Mixup \cite{zhang2018mixup} was first proposed in the field of vision as a simple and effective data augmentation method, which generates new samples by linearly interpolating the two input images. Following this idea, there appears to be a number of mixup-based methods \cite{verma2019manifold,li2021feature,li2022openmixup,cao2024survey,jin2024survey}. CutMix \cite{yun2019cutmix} and Manifold Mixup \cite{verma2019manifold} improved mixup into cutting-based and feature-based. DiffuseMix \cite{islam2024diffusemix} combined a generative model and the mixup method. Beyond the visual domain, mixup has also demonstrated capabilities in tasks such as graph learning \cite{han2022g,jia2024graph} and natural language processing \cite{zhang2020seqmix,zhang2022treemix}. In bioinformatics, MixingDTA \cite{kim2025mixingdta} interpolated embeddings of neighboring entities to improve affinity prediction. R-Mixup \cite{kan2023r} leveraged the log Euclidean distance metrics from the Riemannian manifold, tailoring for biological networks. Due to the complexity of protein structures, existing mixup methods cannot be directly applied to PRL. Our experiments in Section \ref{sec:experiments} also demonstrate that directly applying manifold-based methods compromises model performance. In this work, we begin with the structural information damage issues caused by existing protein DA methods and propose restoration mixing methods. Furthermore, unlike previous approaches that employ DA to construct contrastive views during pre-training \cite{hermosilla2022contrastive,zhangprotein}, we focus on directly utilizing augmented data to enhance performance on downstream tasks, achieving satisfactory results.

\section{Defect of Existing DA Methods}
\label{sec:defect}

\subsection{Protein 1D Sequence and 3D Structure}

A protein $\mathcal{P}$ is expressed as a relational graph $\mathcal{G}_{\mathcal{P}}$, made up of ( $\mathcal{V}, \mathcal{E}, \mathcal{R}$ ). $\mathcal{V}$ is the set of nodes, and each node represents a residue in protein and includes the amino acid residue type and 3D coordinate \cite{lee2023pre}. $\mathcal{E}$ is the set of edges among nodes with their types $\mathcal{R}$, such as the edges between two residues located within a certain distance on the protein sequence or 3D coordinates. When only the residue sequence of the protein is considered (disregarding spatial relationships), the graph reduces to a sequence where edges are solely defined by adjacency in the primary structure (i.e., $\mathcal{R}$ is restricted to peptide-bond connections between consecutive residues, with 3D coordinate information excluded from node attributes). The goal of protein representation learning is to map $\mathcal{G}_{\mathcal{P}}$ (or its sequence-only variant) into a low-dimensional embedding that contains biological information for downstream tasks.

\subsection{Data Augmentation for PRL}
Mainstream protein DA applies specific perturbations or transformations on the original $\mathcal{G}_{\mathcal{P}}$ to generate augmented data $\mathcal{G}_{\mathcal{P}a}$, which can be formulated as $ \mathcal{G}_{\mathcal{P}a} = \operatorname{Aug}\left(\mathcal{G}_{\mathcal{P}}\right)$. We summary commonly used protein DA methods in Table \ref{tab:existing_aug}. The details of each augmentation operation are presented in Appendix \ref{sec:operations}. Note that in this work, we focus on applying augmented data to downstream tasks. Self-supervised pre-training based on augmented views \cite{lee2023pre,zhangprotein} are not involved our discussion and are left for future work. 

We can observe that most 1D operations disrupt the correspondence between 1D and 3D structures, rendering augmented data unusable for methods requiring both 1D and 3D information as input. Typically, these operations are only applicable to purely sequential methods like LSTMs and Transformers. A few operators, such as Mask and Substitute, can preserve this correspondence. 3D operations primarily involve perturbations of coordinates and structure, with such methods typically preserving the sequence, number, and type of residues. The final category of methods focuses on sampling, operating simultaneously on both the 1D sequence and the 3D structure. All augmentation methods result in some degree of loss of the original protein information. However, unlike image augmentation, where altering numerous pixels can still preserve the original meaning, even minor changes to a few amino acids or structural elements in a protein can directly impact its ultimate function. Next, we will validate our hypothesis.

\subsection{Empirical Results}

We select two representative PRL backbone networks, GearNet \cite{zhangprotein} and CDConv \cite{fan2023continuous}, and evaluate the performance from different augmentation methods across four categories of downstream tasks. These two backbones have been widely adopted in many works \cite{quan2024clustering,wang2024multi,cai2024pretrainable}. Experimental details are provided in Section \ref{sec:experimental_setup}. To ensure a fair comparison, we add augmented data as additional samples to the training process. The results are presented in Table \ref{tab:defect}. Noted that among the operations for 1D sequences, we only adopted Substitution and Mask, as the remaining would prevent 1D and 3D structures from aligning. Rotation and Translation were not employed because these augmentations are ineffective for models satisfying SE(3) invariance \cite{zhangprotein,fan2023continuous,jinglearning}. It should be noted that the CDConv uses Gaussian Coordinate Noise by default \cite{fan2023continuous}. 

From the table, we can observe that only a few augmentation methods achieved marginal improvements on a handful of tasks. In most cases, training models with additional augmented data resulted in a significant decline in prediction accuracy. For example, after adding augmented data with replaced residues, the classification accuracy of GearNet and CDCcnv decreased by 3.5\% and 6.2\%, respectively, on Fold. Substructure Sampling and Subspace Sampling resulted in significant performance degradation for both models across all three domains in the Gene Ontology Term Prediction task. The results spanning multiple augmentation operations and prediction tasks on representative PRL backbones validate our hypothesis. Existing augmentation methods disrupt the original protein structure, preventing models from learning accurate protein representations. When applied to downstream tasks, they struggle to deliver performance gains and usually result in degradation.

\section{Methodology}
\begin{figure}[!t]
  \centering
  \includegraphics[scale=0.62]{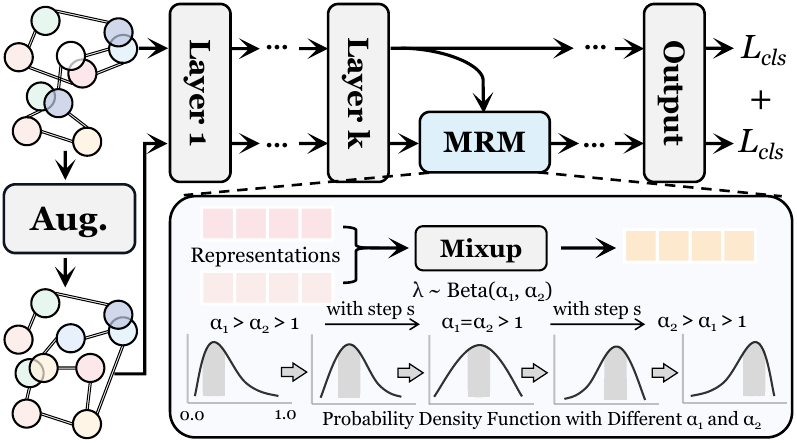}
  \vspace{-0.5em}
  \caption{The illustration of our proposed MRM. The original data is augmented through a randomly sampled operation from the pool. Then, representations of the original data and augmented data are mixed at layer $k$ of the network. Finally, they are used to compute the classification loss.}
  \label{fig:framework}
  \vspace{-1em}
\end{figure}

We illustrate our propose MRM in Figure \ref{fig:framework}. We first elaborate on how MRM performs the mixing operation in Section \ref{sec:mixing}. Then, Section \ref{sec:difficulty} introduces the difficulty scheduler to adjust the beta distribution of MRM during training, providing samples that progress from easy to hard. Finally, the two-stage regularized training strategy involved the original and the mixed representations are presented in Section \ref{sec:two-stage}.

\subsection{Manifold Restore Mixing}
\label{sec:mixing}

The aforementioned analysis and empirical studies raise a question: how can we restore the structural information lost during data augmentation? A straightforward approach is to combine the original data with the augmented data, ensuring the final samples retain both the rich variations from augmentation and the essential structural information from the original data. However, the data format of amino acids and coordinates prevents direct combination.

To achieve this ``addition'', we draw inspiration from manifold mixup \cite{verma2019manifold}. We fuse the original and augmented data at the low-dimensional representation level, enabling the final representation to embody information from both simultaneously. Specifically, given a PRL backbone neural network $Net(\mathcal{G}_{\mathcal{P}}) =Net_k\left(R_k(\mathcal{G}_{\mathcal{P}}\right)$ where $R_k$ denotes the part of the neural network mapping the input data to the hidden representation at layer $k$, and $Net_k$ denotes the part mapping such hidden representation to the output $y$. We first select a random layer $k$ from a set of eligible layers in the network. Then, we process the original data  $\mathcal{G}_{\mathcal{P}}$ and augmented data $\mathcal{G}_{\mathcal{P}a}$ as usual, until reaching layer $k$. This provides us with two intermediate representations $R_k(\mathcal{G}_{\mathcal{P}})$ and $R_k(\mathcal{G}_{\mathcal{P}a})$. Next, we perform mixup \cite{zhang2018mixup} on these representations, producing the mixed representation:
\begin{equation}
\tilde{R_k}=\lambda \cdot R_k(\mathcal{G}_{\mathcal{P}a})+(1-\lambda) \cdot R_k(\mathcal{G}_{\mathcal{P}}), 
\label{eq:mix_representation}
\end{equation}
where $\lambda \sim \operatorname{Beta}(\alpha_1, \alpha_2)$ is the mixup weight from beta distribution. we continue the forward pass in the network from layer $k$ using mixed representation $\tilde{R_k}$ to obtain the final output $\tilde{R_f}$. This output is used to compute the loss value with the original label $y$ and gradients that update the parameters of the neural network. We do not perform any operations on the label $y$. Details of applying mixed representations to training are presented in Section \ref{sec:two-stage}.

\subsection{Difficulty Scheduler}
\label{sec:difficulty}
General mixup sample the weights from a fixed beta distribution with $\alpha_1 = \alpha_2 < 1$ \cite{zhang2018mixup}, which brings two issues.

\noindent \textbf{Small $\alpha$ causes monopoly effect.} First, small alpha (generally 0.2 or 0.3 in previous works\cite{zhang2018mixup,yun2019cutmix,lim2021noisy}) values cause the beta distribution to exhibit peaks at $\lambda=0$ and $\lambda=1$, meaning the mixed samples are either dominated by the original samples or by the augmented samples. The former fails to introduce diverse variations, while the latter yields minimal restoration effects. Inspired by RegMixup \cite{pinto2022using}, we set both $\alpha_1$ and $\alpha_1$ to values greater than 1 in our manifold restore mixing, which brings $\lambda$ closer to 0.5, enabling the mixed representations to incorporate sufficient characteristics from both the original and augmented samples.

\vspace{0.3em}

\noindent \textbf{Fixed $\alpha$ limits the interpolation space.} Secondly, in protein representation learning, existing data usually originates from proteins shaped by natural selection with stable structure and function, providing models with clear fitting targets and lower learning difficulty \cite{watson2023novo,kim2025easy}. In contrast, augmented protein data introduces variations that mimic natural environments \cite{kawabata1999protein,studer2013residue}. Consequently, while maintaining the same labels $y$, this data presents models with higher learning difficulty \cite{hermosilla2022contrastive,sun2024enhancing}. However, fixing $\alpha$ keeps the weight distribution of $\lambda$ unchanged during training, i.e., the sample difficulty remains constant. Existing research has demonstrated that training samples arranged from easy to difficult can effectively enhance model performance and robustness \cite{bengio2009curriculum,wang2021survey}. Therefore, we design a simple  curriculum learning module that provides samples from easy to hard by adjusting the beta distribution during training. Based on the solution of the first issue, given initial value $\alpha_1>\alpha_2>1$, as shown in the lower half of Figure \ref{fig:framework}, we reset the beta distribution every epoch during training with the step $s$:
\begin{equation}
\alpha_1^n = \alpha_1^{n-1} - s, \quad \alpha_2^n = \alpha_2^{n-1} + s.
\label{eq:reset_alpha}
\end{equation}
where $\alpha_1^{n-1}$ and $\alpha_2^{n-1}$ are previous value. The $\alpha_1^{n}$ and $\alpha_2^{n}$ are new values for the beta distribution. So $\lambda \sim \operatorname{Beta}(\alpha_1^{n}, \alpha_2^{n})$ is the new mixup weight. When $\alpha_1$ and $\alpha_2$ are different, the weight distribution is shifted to one side. The larger the difference, the larger the shift. Therefore, we only need to adjust the $\alpha$ value to adjust the distribution of the mixup weight $\lambda$, and thus the sample difficulty. The above adjustment enables the probability distribution of $\lambda$ to gradually shift from a leftward bias to a rightward bias between the intervals $(0,1)$, i.e., increasing the weight of the augmented data in final representations. As the weights increase, the samples gradually move from easier to harder, further improving the performance.

In practice, to maintain simplicity and avoid extreme weight values caused by excessive differences between $\alpha_1$ and $\alpha_2$, we propose an alternative approach to implementing this scheduler. Given initial values $\alpha_1>\alpha_2>1$, we set the scheduling endpoints to $\alpha_2>\alpha_1>1$. That is, $\alpha_1$ gradually increases to $\alpha_2$, while $\alpha_2$ progressively decreases to $\alpha_1$. At this point, the step size $s$ can be calculated based on the total number of epoch $E$, i.e., $s=(\alpha_1-\alpha_2)/E$.

\subsection{Two-Stage Regularized Training}
\label{sec:two-stage}
During training, we constructed an augmentation operation pool containing all augmentation operations listed in Table 5. Given an original protein data $\mathcal{G}_{\mathcal{P}}$, we randomly select an operation from the augmentation pool to augment it, yielding $\mathcal{G}_{\mathcal{P}a}$. The $\mathcal{G}_{\mathcal{P}a}$ is then fed into manifold restore mixing to obtain the final representation $\tilde{R_f}$ as described in Section \ref{sec:mixing}. Next, we present the training process.

At the very beginning of training, the model parameters are merely initialized and have not undergone sufficient learning and optimization. Therefore, to avoid noise and convergence issues caused by mixing representations at this stage, we adopt a two-stage training strategy. In the first stage, we only utilize original data for model training and disable the MRM:
\begin{equation}
\mathcal{L}_{first} = \mathcal{L}_{CLS}\left(Net(\mathcal{G}_{\mathcal{P}}), y\right),
\label{eq:first}
\end{equation}
where $\mathcal{L}_{CLS}$ is the classification loss. Following previous works \cite{fan2023continuous,wangenhancing}, we use negative log-likelihood loss for single-label protein classification tasks and binary cross-entropy loss for multi-label protein classification tasks. After the model parameters converge on the original data, we activate the manifold restore mixing module with the difficulty scheduler in the second stage. Instead of directly replacing the original data and its representation with the mixed representation, we add a term to the Equation (\ref{eq:first}):
\begin{equation}
\mathcal{L}_{second} = \mathcal{L}_{CLS}\left(Net(\mathcal{G}_{\mathcal{P}}), y\right)+  \gamma\mathcal{L}_{CLS}(\tilde{R_f}, y),
\label{eq:second}
\end{equation}
where $\gamma$ is the weight used to control the contribution of the mixed sample loss. In Equation (\ref{eq:second}), we explicitly combine empirical risk minimization (ERM, the first term) based on the original protein data with vicinal risk minimization (VRM, the second term) based on the mixed samples, which has been proven to enhance model performance effectively \cite{pinto2022using}. The former ensures in-distribution fitting accuracy, while the latter explores reasonable variation within the sample vicinity based on the restored representation, thereby avoiding the limitations of a single paradigm.

\section{Experiments}
\label{sec:experiments}

\begin{table*}[!t]
  \centering
  \caption{Performance comparison of different PRL methods and our MRM. The ``w/ MRM'' represents adding our MRM on corresponding PRL methods. The performance of baseline methods are copied from \cite{fan2023continuous}. We report the maximum of F1 score ($F_{max}$) for Enzyme Commission and Gene Ontology prediction tasks; and Top-1 accuracy (\%) for Fold and Reaction classification.}
  \vspace{-1em}
  \renewcommand\arraystretch{0.9}
  \scalebox{1.0}{
    \begin{tabular}{c|l|cccc|c|ccc|c}
    \toprule
    \multicolumn{2}{c|}{\multirow{2}[2]{*}{Method}} & \multicolumn{4}{c|}{Fold Classification} & Enzyme & \multicolumn{3}{c|}{Gene Ontology} & Enzyme \\
    \multicolumn{2}{c|}{} & Fold  & Superfamily & Family & Average & Reaction & BP & MF & CC & Commission \\
    \midrule
    \multirow{4}[2]{*}{1D} & CNN \cite{shanehsazzadeh2020transfer} & 11.3  & 13.4  & 53.4  & 26.0  & 51.7  & 0.244  & 0.354  & 0.287  & 0.545  \\
          & ResNet \cite{rao2019evaluating}  & 10.1  & 7.21  & 23.5  & 13.6  & 24.1  & 0.280  & 0.405  & 0.304  & 0.605  \\
          & LSTM  \cite{rao2019evaluating} & 6.41  & 4.33  & 18.1  & 9.6   & 11.0  & 0.225  & 0.321  & 0.283  & 0.425  \\
          & Transformer \cite{shanehsazzadeh2020transfer} & 9.22  & 8.81  & 40.4  & 19.5  & 26.6  & 0.264  & 0.211  & 0.405  & 0.238  \\
    \midrule
    \multirow{3}[2]{*}{3D} & GCN \cite{kipf2016semi} & 16.8  & 21.3  & 82.8  & 40.3  & 67.3  & 0.252  & 0.195  & 0.329  & 0.320  \\
          & GAT \cite{velivckovic2018graph} & 12.4  & 16.5  & 72.7  & 33.9  & 55.6  & 0.284  & 0.317  & 0.385  & 0.368  \\
          & 3DCNN \cite{derevyanko2018deep} & 31.6  & 45.4  & 92.5  & 56.5  & 72.2  & 0.240  & 0.147  & 0.305  & 0.077  \\
    \midrule
    \multirow{12}[10]{*}{(3+1)D} & GraphQA \cite{baldassarre2021graphqa} & 23.7  & 32.5  & 84.4  & 46.9  & 60.8  & 0.308  & 0.329  & 0.413  & 0.509  \\
          & GVP  \cite{jinglearning} & 16.0  & 22.5  & 83.8  & 40.8  & 65.5  & 0.326  & 0.426  & 0.420  & 0.489  \\
          & IEConv \cite{hermosillaintrinsic} & 45.0  & 69.7  & 98.8  & 71.2  & 87.2  & -     & -     & -     & - \\
          & New IEConv \cite{hermosilla2022contrastive} & 47.6  & 70.2  & 99.2  & 72.3  & 87.2  & 0.374  & 0.544  & 0.444  & 0.735  \\
\cmidrule{2-11} & GearNet \cite{zhangprotein} & 28.4  & 42.6  & 95.3  & 55.4  & 79.4  & 0.356  & 0.503  & 0.414  & 0.730  \\
          \rowcolor{mygray}\cellcolor{white} & \textbf{w/ MRM} & \textbf{31.1} & \textbf{45.5} & \textbf{95.8} & \textbf{57.5} & \textbf{81.1} & \textbf{0.368} & \textbf{0.514} & \textbf{0.426} & \textbf{0.751} \\
\cmidrule{2-11} & GearNet-Edge  \cite{zhangprotein} & 44.0  & 66.7  & 99.1  & 69.9  & 86.6  & 0.403  & 0.580  & 0.450  & 0.810  \\
          \rowcolor{mygray}\cellcolor{white} & \textbf{w/ MRM} & \textbf{47.1} & \textbf{69.9} & \textbf{99.3} & \textbf{72.1} & \textbf{87.5} & \textbf{0.412} & \textbf{0.591} & \textbf{0.463} & \textbf{0.829} \\
\cmidrule{2-11} & ProNet \cite{wanglearning} & 51.5  & 69.9  & 99.0  & 73.5  & 86.0  & -     & -     & -     & - \\
          \rowcolor{mygray}\cellcolor{white} & \textbf{w/ MRM} & \textbf{53.7} & \textbf{72.7} & \textbf{99.3} & \textbf{75.2} & \textbf{87.2} & -     & -     & -     & - \\
\cmidrule{2-11} & CDConv \cite{fan2023continuous} & 56.7  & 77.7  & 99.6  & 78.0  & 88.5  & 0.453  & 0.654  & 0.479  & 0.843  \\
          \rowcolor{mygray}\cellcolor{white} & \textbf{w/ MRM} & \textbf{60.8} & \textbf{81.7} & \textbf{99.7} & \textbf{80.7} & \textbf{89.5} & \textbf{0.462} & \textbf{0.667} & \textbf{0.496} & \textbf{0.868} \\
    \bottomrule
    \end{tabular}}%
  \label{tab:main}%
\end{table*}%

\begin{table*}[!t]
  \centering
  \caption{Performance comparison of different mixup-based baselines and our MRM.}
  \vspace{-1em}
  \renewcommand\arraystretch{0.9}
  \scalebox{1.0}{
    \begin{tabular}{c|l|cccc|c|ccc|c}
    \toprule
    \multicolumn{2}{c|}{\multirow{2}[2]{*}{Method}} & \multicolumn{4}{c|}{Fold Classification} & Enzyme & \multicolumn{3}{c|}{Gene Ontology} & Enzyme \\
    \multicolumn{2}{c|}{} & Fold  & Superfamily & Family & Average & Reaction & BP    & MF    & CC    & Commission \\
    \midrule
    \multirow{8}[3]{*}{GearNet} & Base  & 28.4  & 42.6  & 95.3  & 55.4  & 79.4  & 0.356  & 0.503  & 0.414  & 0.730  \\
\cmidrule{2-11}          & w/ Manifold & 0.1   & 12.8  & 0.2   & 4.4   & 77.8  & 0.349  & 0.474  & 0.386  & 0.715  \\
          & w/ TS Manifold & 27.5  & 41.3  & 94.2  & 54.3  & 79.0  & 0.353  & 0.491  & 0.407  & 0.733  \\
          & w/ NFM & 0.1   & 10.4  & 0.1   & 3.5   & 77.5  & 0.343  & 0.487  & 0.371  & 0.694  \\
          & w/ TS NFM & 28.0  & 41.6  & 94.7  & 54.8  & 79.2  & 0.348  & 0.494  & 0.403  & 0.711  \\
          & w/ RegManifold & 28.2  & 42.1  & 95.0  & 55.1  & 78.9  & 0.352  & 0.492  & 0.396  & 0.735  \\
          & w/ TS RegManifold & 28.8  & 43.1  & 95.4  & 55.8  & 79.7  & 0.355  & 0.506  & 0.418  & 0.740  \\
          \rowcolor{mygray}\cellcolor{white} & \textbf{w/ MRM} & \textbf{31.1} & \textbf{45.5} & \textbf{95.8} & \textbf{57.5} & \textbf{81.1} & \textbf{0.368} & \textbf{0.514} & \textbf{0.426} & \textbf{0.751} \\
    \midrule
    \multirow{8}[3]{*}{CDConv} & Base  & 56.7  & 77.7  & 99.6  & 78.0  & 88.5  & 0.453  & 0.654  & 0.479  & 0.843  \\
    & w/ Manifold & 0.1   & 25.0  & 0.1   & 8.4   & 86.1  & 0.437  & 0.627  & 0.437  & 0.845  \\
          & w/ TS Manifold & 55.8  & 77.6  & 99.3  & 77.6  & 87.9  & 0.445  & 0.641  & 0.462  & 0.847  \\
          & w/ NFM & 0.3   & 28.4  & 0.6   & 9.8   & 87.4  & 0.435  & 0.608  & 0.458  & 0.823  \\
          & w/ TS NFM & 56.9  & 78.1  & 99.5  & 78.2  & 88.3  & 0.439  & 0.635  & 0.466  & 0.835  \\
          & w/ RegManifold & 55.7  & 77.0  & 99.4  & 77.4  & 87.8  & 0.449  & 0.640  & 0.473  & 0.845  \\
          & w/ TS RegManifold & 57.2  & 78.6  & 99.2  & 78.3  & 88.6  & 0.455  & 0.649  & 0.483  & 0.852  \\
          \rowcolor{mygray}\cellcolor{white} & \textbf{w/ MRM} & \textbf{60.8} & \textbf{81.7} & \textbf{99.7} & \textbf{80.7} & \textbf{89.5} & \textbf{0.462} & \textbf{0.667} & \textbf{0.496} & \textbf{0.868} \\
    \bottomrule
    \end{tabular}}%
  \label{tab:mixup-based}%
  \vspace{-1.0em}
\end{table*}%

\subsection{Experimental Setup}
\label{sec:experimental_setup}

\noindent \textbf{Evaluation Tasks and Datasets.}
Following the previous work \cite{fan2023continuous,wangenhancing}, we evaluate the effectiveness of our approach across four protein-related tasks: Protein Fold Classification (FOLD), Enzyme Reaction Classification (ER), Gene Ontology Term Prediction (GO), and Enzyme Commission (EC) number prediction. For FOLD, we evaluate performance under three scenarios: fold, superfamily, and family classification. For GO, we assess performance across three sub-tasks: biological process (BP), molecular function (MF), and cellular component (CC) ontology term prediction. We conduct five runs and report the average results for our methods. Details of tasks and datasets are provided in the Appendix \ref{sec:tasks}.

\vspace{0.3em}

\noindent \textbf{Evaluation Metrics.}
The classification accuracy is used for single-label classification tasks (FOLD and Reaction). For multi-label classification tasks, GO and EC, we adopt the protein-centric maximum F-score $F_{max}$, which is based on the precision and recall of the predictions.

\vspace{0.3em}

\noindent \textbf{Baseline Methods.} To comprehensively evaluate the superiority of our approach, we selected various PRL methods and mixup methods as baselines. The PRL method can be categorized into three categories based on its inputs, which could be 1D sequences, 3D structures, or both. 1) Sequence-based encoders: \textbf{CNN} \cite{shanehsazzadeh2020transfer}, \textbf{ResNet} \cite{rao2019evaluating}, \textbf{LSTM} \cite{rao2019evaluating} and \textbf{Transformer} \cite{rao2019evaluating}. 2) Structure-based methods: \textbf{GCN} \cite{kipf2016semi}, \textbf{GAT} \cite{velivckovic2018graph}, \textbf{3DCNN} \cite{derevyanko2018deep}. 3) Sequence-structure based methods: \textbf{GraphQA} \cite{baldassarre2021graphqa}, \textbf{GVP} \cite{jinglearning}, \textbf{IEConv} \cite{hermosillaintrinsic}, \textbf{New IEConv} \cite{hermosilla2022contrastive}, \textbf{GearNet} \cite{zhangprotein}, \textbf{GearNet-edge} \cite{zhangprotein}, \textbf{ProNet} \cite{wanglearning}, \textbf{CDConv} \cite{fan2023continuous}. Our research focuses on methods that consider both 1D and 3D structures. Therefore, among these methods, we selected GearNet, GearNet-edge, ProNet, and CDConv as the backbones to implement our method. These methods are selected not only for their representativeness and SOTA performance, but also because they provide complete open-source code with implementation details, which facilitates rigorous and fair comparisons. Since there are currently no mixup methods specifically designed for PRL, and protein data cannot be directly mixed like images, we compared several general and representative manifold-based mixup methods, including \textbf{Manifold Mixup} \cite{verma2019manifold}, \textbf{NFM} \cite{lim2021noisy}, and \textbf{RegManifold}. Since RegMixup \cite{pinto2022using} can only be applied to images, we combined it with Manifold Mixup \cite{verma2019manifold} to form RegManifold.

\begin{table*}[!t]
  \centering
  \caption{Performance comparison between different protein language models and our method across four tasks. Baseline results are from \cite{wangenhancing}, \cite{wangproteinadapter} and \cite{wang2025aligning}. The ``AR'' represents the average ranking a method achieves across all evaluated tasks.}
    \vspace{-1em}
    \renewcommand\arraystretch{0.9}
  \scalebox{1.0}{
    \begin{tabular}{l|cc|cccc|c|ccc|c|c}
    \toprule
    \multicolumn{1}{c|}{\multirow{2}[2]{*}{Method}} & \multicolumn{2}{c|}{\multirow{2}[2]{*}{Pretraining Dataset}} & \multicolumn{4}{c|}{Fold Classification} & \multirow{2}[2]{*}{ER} & \multicolumn{3}{c|}{Gene Ontology} & \multirow{2}[2]{*}{EC} & \multirow{2}[2]{*}{AR} \\
          & \multicolumn{2}{c|}{} & Fold  & Super. & Family & Ave.  &       & BP    & MF    & CC    &       &  \\
    \midrule
    DeepFRI \cite{ozturk2018deepdta} & \multicolumn{1}{r}{Pfam} & \multicolumn{1}{l|}{10M} & 15.3  & 20.6  & 73.2  & 36.4  & 63.3  & 0.399  & 0.465  & 0.460  & 0.631  & 11.6 \\
    ESM-1b \cite{rives2021biological} & \multicolumn{1}{r}{UniRef50} & \multicolumn{1}{l|}{24M} & 26.8  & 60.1  & 97.8  & 61.6  & 83.1  & 0.470  & 0.657  & 0.488  & 0.864  & 7.1 \\
    ProtBERT-BFD \cite{elnaggar2021prottrans} & \multicolumn{1}{r}{BFD} & \multicolumn{1}{l|}{2.1B} & 26.6  & 55.8  & 97.6  & 60.0  & 72.2  & 0.279  & 0.456  & 0.408  & 0.838  & 11.3 \\
    IEConv (Residue Level) \cite{hermosilla2022contrastive} & \multicolumn{1}{r}{PDB} & \multicolumn{1}{l|}{476K} & 50.3  & 80.6  & 99.7  & 76.9  & 88.1  & 0.468  & 0.661  & 0.516  & -     & 3.3 \\
    LM-GVP \cite{wang2022lm} & \multicolumn{1}{r}{UniRef100} & \multicolumn{1}{l|}{0.21B} & -     & -     & -     & -     & -     & 0.417  & 0.545  & 0.527  & 0.664  & 8.5 \\
    ESM-2 \cite{lin2023evolutionary} & \multicolumn{1}{r}{UniRef50} & \multicolumn{1}{l|}{24M} & -     & 78.9  & 99.9  & -     & 87.2  & 0.460  & 0.661  & 0.445  & 0.880  & 5.6 \\
    GearNet-E-IE with RTP \cite{zhangprotein} & \multicolumn{1}{r}{AlphaFoldDB} & \multicolumn{1}{l|}{805K} & 48.8  & 71.0  & 99.4  & 73.1  & 86.6  & 0.430  & 0.604  & 0.465  & 0.843  & 8.6 \\
    GearNet-E-IE with DP \cite{zhangprotein} & \multicolumn{1}{r}{AlphaFoldDB} & \multicolumn{1}{l|}{805K} & 50.9  & 73.5  & 99.4  & 74.6  & 87.5  & 0.448  & 0.616  & 0.464  & 0.839  & 7.7 \\
    GearNet-E-IE with AP \cite{zhangprotein} & \multicolumn{1}{r}{AlphaFoldDB} & \multicolumn{1}{l|}{805K} & 56.5  & 76.3  & 99.6  & 77.5  & 86.8  & 0.458  & 0.625  & 0.473  & 0.853  & 5.9 \\
    GearNet-E-IE with DP \cite{zhangprotein} & \multicolumn{1}{r}{AlphaFoldDB} & \multicolumn{1}{l|}{805K} & 51.8  & 77.8  & 99.6  & 76.4  & 87.0  & 0.458  & 0.626  & 0.465  & 0.859  & 6.1 \\
    GearNet-E-IE with MC \cite{zhangprotein} & \multicolumn{1}{r}{AlphaFoldDB} & \multicolumn{1}{l|}{805K} & 54.1  & 80.5  & 99.9  & 78.2  & 87.5  & 0.490  & 0.654  & 0.488  & 0.874  & 3.0  \\
    SaProt-PDB \cite{susaprot} & \multicolumn{1}{r}{PDB} & \multicolumn{1}{l|}{60K} & 52.5  & 77.8  & 99.6  & 76.6  & 87.7  & 0.465  & 0.669  & 0.415  & 0.888  & 4.4 \\
    \rowcolor{mygray} \textbf{CDConv w/ MRM} & \multicolumn{2}{c|}{-} & \textbf{60.8} & \textbf{81.7} & 99.7  & \textbf{80.7} & \textbf{89.5} & 0.462  & 0.667 & 0.496  & 0.868  & \textbf{2.3} \\
    \bottomrule
    \end{tabular}}%
  \label{tab:protein_llm}%
\end{table*}%

\begin{table*}[!t]
  \centering
  \caption{Performance comparison of different methods on GO and EC under different sequence cutoffs.}
      \vspace{-1.0em}
      \setlength{\tabcolsep}{1mm}{
  \scalebox{0.85}{
    \begin{tabular}{l|ccccc|ccccc|ccccc|ccccc}
    \toprule
    \multicolumn{1}{c|}{\multirow{2}[2]{*}{Method}} & \multicolumn{5}{c|}{GO-BP}  & \multicolumn{5}{c|}{GO-MF}  & \multicolumn{5}{c|}{GO-CC}  & \multicolumn{5}{c}{EC} \\
          & 30\%  & 40\%  & 50\%  & 70\%  & 95\%  & 30\%  & 40\%  & 50\%  & 70\%  & 95\%  & 30\%  & 40\%  & 50\%  & 70\%  & 95\%  & 0.300  & 0.400  & 0.500  & 0.700  & 0.950  \\
    \midrule
    CNN   & 0.197  & 0.195  & 0.197  & 0.211  & 0.244  & 0.238  & 0.243  & 0.256  & 0.292  & 0.354  & 0.258  & 0.257  & 0.260  & 0.263  & 0.387  & 0.366  & 0.361  & 0.372  & 0.429  & 0.545  \\
    ResNet & 0.230  & 0.230  & 0.234  & 0.249  & 0.280  & 0.282  & 0.288  & 0.308  & 0.347  & 0.405  & 0.277  & 0.273  & 0.280  & 0.278  & 0.304  & 0.409  & 0.412  & 0.450  & 0.526  & 0.605  \\
    LSTM  & 0.194  & 0.192  & 0.195  & 0.205  & 0.225  & 0.223  & 0.229  & 0.245  & 0.276  & 0.321  & 0.263  & 0.264  & 0.269  & 0.270  & 0.283  & 0.249  & 0.249  & 0.270  & 0.333  & 0.425  \\
    Transformer & 0.267  & 0.265  & 0.262  & 0.262  & 0.264  & 0.184  & 0.187  & 0.195  & 0.204  & 0.211  & 0.378  & 0.382  & 0.388  & 0.395  & 0.405  & 0.167  & 0.173  & 0.175  & 0.197  & 0.238  \\
    GearNet & 0.251  & 0.250  & 0.248  & 0.248  & 0.252  & 0.180  & 0.183  & 0.187  & 0.194  & 0.195  & 0.318  & 0.318  & 0.320  & 0.323  & 0.329  & 0.245  & 0.246  & 0.246  & 0.280  & 0.320  \\
    GearNet-E & 0.345  & 0.347  & 0.354  & 0.378  & 0.403  & 0.444  & 0.461  & 0.490  & 0.537  & 0.580  & 0.394  & 0.394  & 0.401  & 0.408  & 0.450  & 0.625  & 0.646  & 0.694  & 0.757  & 0.810  \\
    CDConv & 0.381  & 0.390  & 0.401  & 0.428  & 0.453  & 0.533  & 0.533  & 0.577  & 0.621  & 0.654  & 0.428  & 0.435  & 0.440  & 0.451  & 0.479  & 0.689  & 0.720  & 0.760  & 0.809  & 0.843  \\
    \midrule
    \rowcolor{mygray} \textbf{Ours} & \textbf{0.389} & \textbf{0.398} & \textbf{0.410} & \textbf{0.439} & \textbf{0.462} & \textbf{0.544} & \textbf{0.563} & \textbf{0.588} & \textbf{0.633} & \textbf{0.667} & \textbf{0.445} & \textbf{0.451} & \textbf{0.456} & \textbf{0.470} & \textbf{0.496} & \textbf{0.725} & \textbf{0.752} & \textbf{0.790} & \textbf{0.835} & \textbf{0.868} \\
    \bottomrule
    \end{tabular}}}%
    \vspace{-0.5em}
  \label{tab:cutoffs}%
\end{table*}%

\subsection{Overall Improvements}

The performance of different PRL backbone methods and adding our proposed MRM are summarized in Table \ref{tab:main}. From the table, we have the following observations:
\begin{itemize}
    \item Overall, methods that model protein 3D structures (e.g., 3DCNN) outperform those that model only 1D residue sequences, while approaches that simultaneously perceive both 1D and 3D structures (e.g., ProNet, CDConv)  achieve satisfactory performance. The 1D and 3D structures provide different information about proteins \cite{hu2024learning}, so understanding both facilitates the learning of accurate protein representations for downstream tasks. 
    \item MRM achieves significant performance improvements across four representative 1D and 3D perception PRL backbone networks. For the Fold Classification task, it delivers absolute performance gains of 3.1\% and 4.1\% for GearNet-Edge and CDConv regarding Fold. For Superfamily, the absolute increases are 2.2\% and 3.0\%, respectively. Regarding Enzyme Reaction Classification, it delivers an absolute improvement equal to or exceeding 1\% for four backbones. Compared to the failure of traditional data augmentation methods on GO Term Prediction, MRM achieved performance improvements in all cases. On the EC prediction task, it improves GearNet and CDConv by 2.1\% and 2.5\%.
    \item The above results demonstrate that MRM enables the learned representations to more accurately and comprehensively reflect protein structures, catalytic activities, functions, and catalytic biochemical reactions. Combining Table \ref{tab:defect} and Table \ref{tab:main}, MRM provides models with data that incorporates both original structural information and diverse variations through its representation-level restore effects and difficulty scheduler. It enables traditional protein DA to be directly applied to improve downstream tasks, showing both effectiveness and generalization.
\end{itemize}

\subsection{Comparison with Mixup-Based Methods}
We compared the proposed method with existing mixup-based methods. The results are presented in Table \ref{tab:mixup-based}. In our experiments, we observed that directly enabling existing methods at the start of training prevents the model from learning meaningful protein representations, resulting in inferior performance. Therefore, we introduced a ``TS'' (Two-Stage) variant for each mixup-based method: first, train the base model; then enable the corresponding augmentation method in the second stage. We can observe that existing manifold mixup and its variants fail to apply to PRL, leading to performance degradation in most cases and even preventing models from learning meaningful representations in some instances. We attribute this phenomenon to the fact that these methods mix different protein data, which may generate biologically implausible samples in the manifold space, thereby preventing the model from learning accurate representations. Thanks to representations thoroughly learned from the original data, we also observe that two-stage training improves performance. Our MRM demonstrates superiority among different tasks and backbones.

\subsection{Comparison with Protein LLMs}
Recently, protein-specific large language models (Protein LLMs) have revolutionized protein science by enabling more efficient structure prediction, function annotation, and design \cite{xiao2025protein}. These methods typically rely on extensive pre-training with massive datasets to learn representations. Here, we compare our method with protein large language models and present the results in Table \ref{tab:protein_llm}. We can observe that even without any pre-training, our method achieves the best performance on Fold and Enzyme Reaction Classification. It also achieves the highest average ranking across all methods. Furthermore, our representational restore paradigm may also offer insights for improving pre-training or self-supervised learning methods, as data augmentation may also required during pre-training to construct self-supervised signals for Protein LLMs \cite{lee2023pre,zhangprotein}.

\begin{table*}[!t]
  \centering
  \caption{The results of ablation study. The PRL backbone here is the CDConv.}
    \vspace{-1em}
    \renewcommand\arraystretch{0.9}
  \scalebox{1.0}{
    \begin{tabular}{l|cccc|c|ccc|c}
    \toprule
    \multicolumn{1}{c|}{\multirow{2}[2]{*}{Variant}} & \multicolumn{4}{c|}{Fold Classification} & Enzyme & \multicolumn{3}{c|}{Gene Ontology} & Enzyme \\
          & Fold  & Superfamily & Family & Average & Reaction & BP    & MF    & CC    & Commission \\
    \midrule
    Ours & 60.8  & 81.7  & 99.7  & 80.7  & 89.5  & 0.462  & 0.667  & 0.492  & 0.868  \\
    \midrule
    (1) w/o 1D Sequence Augmentation & 60.0  & 81.2  & 99.7  & 80.3  & 89.2  & 0.458  & 0.662  & 0.484  & 0.859  \\
    (2) w/o 3D Structure Augmentation & 59.7  & 80.5  & 99.6  & 79.9  & 89.1  & 0.460  & 0.660  & 0.487  & 0.857  \\
    (3) w/o Sampling Augmentation & 58.9  & 80.1  & 99.6  & 79.5  & 88.9  & 0.458  & 0.658  & 0.486  & 0.861  \\
    (4) w/o Two-Stage Training & 59.2  & 79.5  & 99.5  & 79.4  & 89.2  & 0.459  & 0.661  & 0.489  & 0.855  \\
    (5) w/o Difficulty Scheduler & 59.6  & 80.6  & 99.6  & 79.9  & 89.0  & 0.457  & 0.662  & 0.490  & 0.860  \\
    \midrule
    Base & 56.7  & 77.7  & 99.6  & 78.0  & 88.5  & 0.453  & 0.654  & 0.479  & 0.843  \\
    \bottomrule
    \end{tabular}}%
  \label{tab:ablation}%
\end{table*}%

\subsection{Go and EC under Different Sequence Cutoffs}
\label{sec:cutoffs}
For gene ontology term prediction and enzyme commission number prediction tasks, we follow the evaluation protocol used by previous work \cite{gligorijevic2021structure,quan2024clustering,wangenhancing} to divide the test set based on the sequence similarity to the training set. The cutoffs are 30\%, 40\%, 50\%, 70\%, and 95\%. This systematic evaluation across different similarity thresholds helps assess model generalization ability and robustness \cite{wangenhancing}. The results are presented in Table \ref{tab:cutoffs}. We observe consistent improvements across different cutoffs and tasks. Even at lower sequence similarity cutoffs, the gains achieved are comparable to those at higher sequence similarity cutoffs. This further demonstrates that our method provides the backbone network with training samples that contain rich information and sufficient variation, thereby enhancing model accuracy and generalization. The effective gains achieved at lower sequence similarity cutoffs also underscore the practical applicability of our approach in real-world scenarios.

\subsection{Ablation Study}
We conduct an ablation study to verify the effectiveness of
different components in our MRM. We compare the following variants: (1): Removing the Substitution and Mask from the operation pool. (2): Removing the Torsion Angle Perturbation and Gaussian Coordinate Noise from the operation pool. (3): Removing the Substructure Sampling and Subspace Sampling from the operation pool. (4): Enabling the MRM (i.e., Equation (\ref{eq:second})) at the beginning of training without a two-stage strategy. (5): Removing the Difficulty Scheduler in Section \ref{sec:difficulty} with a fixed beta distribution. The results are presented in Table \ref{tab:ablation}. We can observe that all components contribute to the final performance. For augmentation operations, the best performance is achieved when alloperations are incorporated. Thanks to our restore module, information loss caused by different operations can be mitigated, while the augmentations themselves provide sufficient variation. Simultaneously, different augmentation operations contribute differently to various tasks. Sampling augmentation contributes more to fold classification, while sequence-based operations show significant benefits for GO and EC. Additionally, the two-stage training strategy and difficulty scheduler help the model learn more accurate and robust representations.

\begin{figure}[!t]
  \centering
  \vspace{-0.5em}
  \includegraphics[scale=0.47]{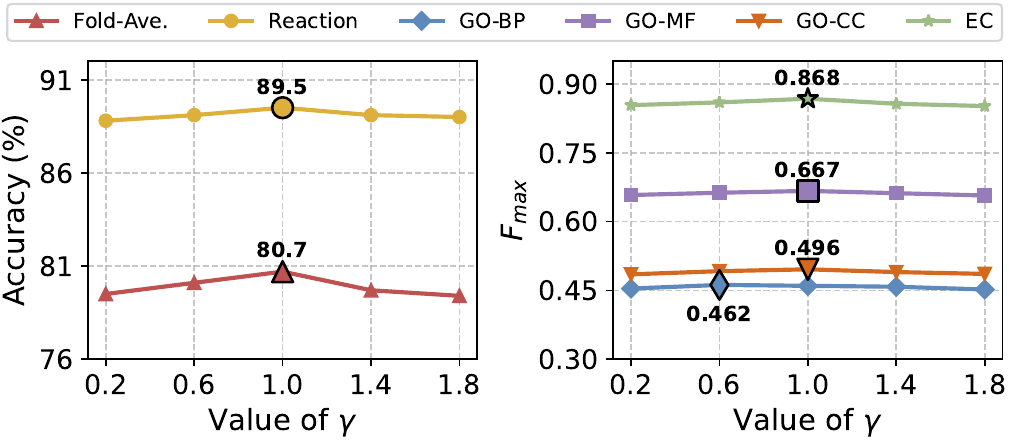}
  \caption{Effect of strength $\gamma$ on model performance. We highlight the best performance of each task for ease of reading.}
  \label{fig:gamma}
  \vspace{-1em}
\end{figure}

\subsection{Hyperparameter Investigation}

We investigate the impact of mixed sample loss strength $\gamma$ on performance. The results are illustrated in Figure \ref{fig:gamma}. The impact of initial values $\alpha_1$ and $\alpha_2$ on performance is investigated in Appendix \ref{sec:more_hyperparameter}. We observe that optimal performance typically occurs when $\gamma=1.0$, i.e., when the original and mixed samples have the same weights in training. This demonstrates that the augmented data after restoration exhibits nearly identical functionality to the original data, enabling the model to learn representations equally well from both, validating the effectiveness of our MRM.

\subsection{Visualization of MRM Restoration}
\begin{figure}[!t]
  \centering
  \includegraphics[scale=0.16]{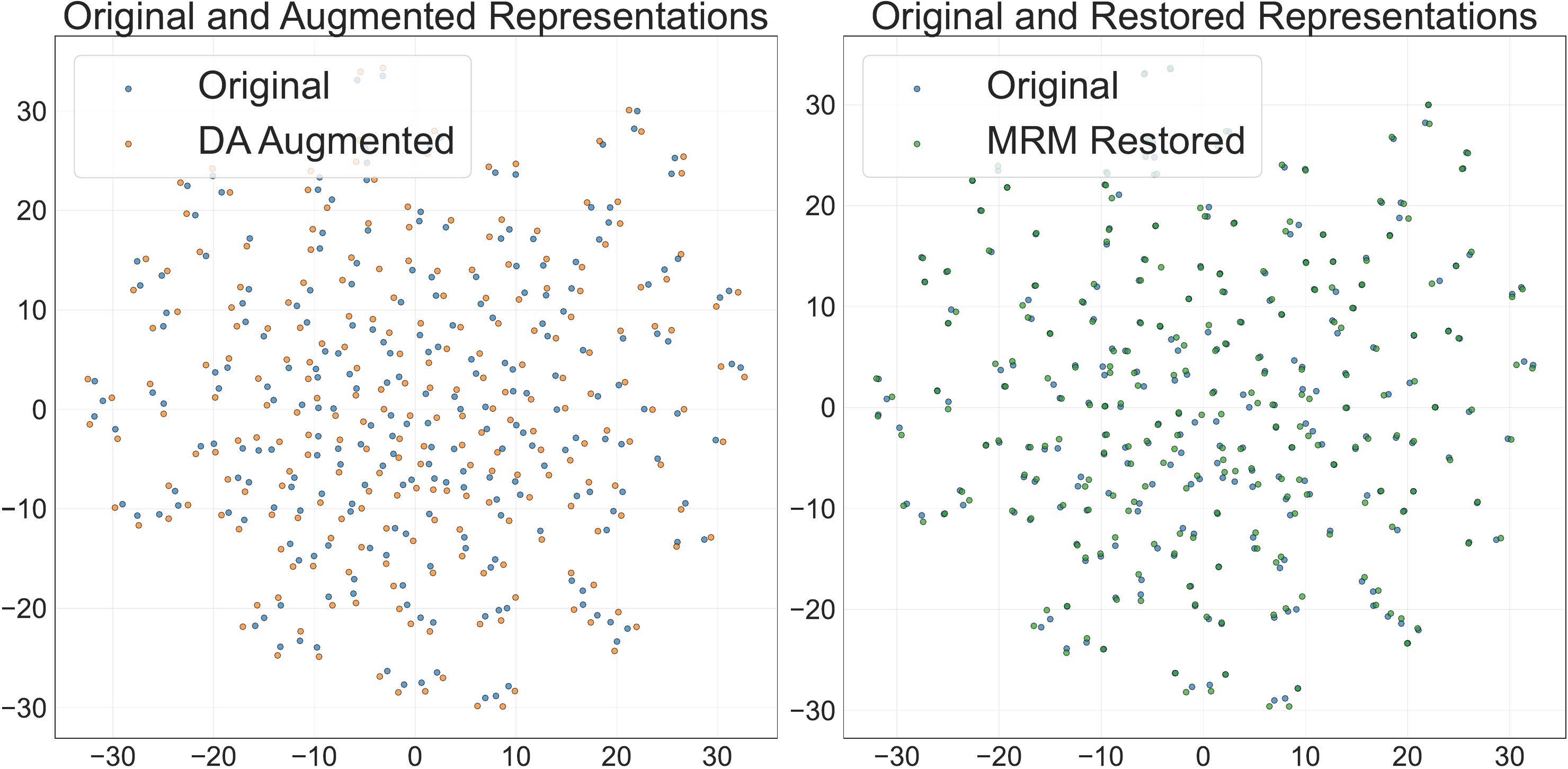}
  \vspace{-0.5em}
  \caption{The t-SNE visualization of different representations in the intermediate layer of the backbone network (CDConv).}
  \label{fig:visualization_1}
  \vspace{-1em}
\end{figure}

We use t-SNE \cite{maaten2008visualizing} to visualize the representations of the augmented data after restoration (i.e., the final mixed representation) and those without restoration in the intermediate layer (Fold task). As illustrated in Figure \ref{fig:visualization_1}, before restoration, the distance between the representation of proteins augmented by DA and the original protein representation is relatively large (average Euclidean distance: 5.67). This distance is reduced after using MRM to restore the representations of augmented data (average Euclidean distance: 4.02), but some variation remained. This restoration process achieves our goal: to provide samples that retain both the original structural information and diverse variations.

\section{Conclusion}
In this work, we analyze and empirically reveal the structure information destruction and performance degradation of existing protein DA operations for PRL. We propose manifold restore mixing to provide augmented data that preserves original structures while incorporating rich variations. We equip MRM with a sample difficulty scheduler with two-stage regularized training, further improving the performance. Comprehensive experiments demonstrate the effectiveness and generalization ability of our proposed MRM. Our work not only promotes the design of DA methods in computational biology but may also offer insights into constructing self-supervised signals for Protein LLM pre-training. For future work, we aim to dynamically adjust operations by integrating Bayesian optimization \cite{wu2019hyperparameter} or reinforcement learning \cite{cubuk2019autoaugment}. Also, extending the MRM to the multimodal PRL \cite{su2024protrek,susaprot} or incorporating structure prediction methods, such as AlphaFold3 \cite{abramson2024accurate}, to further optimize the DA process are promising avenues for exploration.

\section{Limitations and Ethical Considerations}
The datasets are publicly available from previous works, downloaded via official APIs. We do not disclose any non-open-source data. We ensure that our actions comply with ethical standards and have no privacy concerns. Due to resource constraints, we are currently unable to evaluate MRM during Protein LLMs pre-training. Furthermore, since the representation restore performs in the latent space, in-depth biological analysis may prove challenging. We evaluated MRM solely through the performance profiles of downstream tasks. Future research should incorporate more biological perspectives (e.g., conservation of active-site residues and 3D structure similarity) to facilitate the development of more effective methods. Also, we clarify that restore refers to restoring structure-relevant latent information that is degraded by data augmentation operations, rather than reconstructing a fully physically realizable 3D conformation in coordinate space.

\begin{acks}
This research is partially supported by the National Natural Science Foundation of China under Grant No. (62576083, 62432003, U25A20431). This research is also supported by the Ministry of Education, Singapore, under its Academic Research Fund (AcRF) Tier 1 grant, and funded through the SMU-SUTD Internal Research Grant Call (SMU-SUTD 2023\_02\_01), and in part by the Ministry of Education, Singapore, under its Academic Research Fund Tier 2 (Award No. MOE-T2EP201230015). The authors greatly appreciate the anonymous reviewers for their valuable comments.
\end{acks}


\bibliographystyle{ACM-Reference-Format}
\balance
\bibliography{references}


\appendix

\section{Overview}
The appendix is organized as follows:
\begin{itemize}
    \item Section \ref{sec:operations} provides details about protein data augmentation operations involved in this work.
    \item Section \ref{sec:experimental_details} provides details about our experimental settings, including evaluation tasks (\ref{sec:tasks}) and implementation details (\ref{sec:implementation_details}).
    \item Section \ref{sec:additional_results} provides additional hyperparameter investigation (\ref{sec:more_hyperparameter}).
\end{itemize}
\renewcommand{\thesection}{\Alph{section}}

\section{Augmentation Operations}
\label{sec:operations}

Here, we provide a detailed explanation of every operation that appears in Section \ref{sec:defect}.

\subsection{1D Sequence Operations}

\noindent \textbf{Deletion} \cite{sun2024enhancing}. It randomly deletes some amino acids from the protein sequence to generate a new sequence variant.

\vspace{0.3em}

\noindent \textbf{Crop} \cite{sun2024enhancing}. It is similar to Deletion but operating on subsequences. It randomly deletes a continuous subsequence from the protein sequence instead of single amino acids.

\vspace{0.3em}

\noindent \textbf{Cut} \cite{sun2024enhancing}. It cuts the protein sequence into multiple sub-sequences and randomly reassembles them to form a new sequence.

\vspace{0.3em}

\noindent \textbf{Subsequence} \cite{sun2024enhancing,zhou2025protclip}. It selects some sub-sequences and assembles them sequentially into a new sequence.

\vspace{0.3em}

\noindent \textbf{Swap} \cite{sun2024enhancing}. It randomly swaps the positions of two amino acids in the protein sequence to introduce minor changes.

\vspace{0.3em}

\noindent \textbf{Shuffle} \cite{sun2024enhancing}. It first randomly selects a subsequence within the protein amino acid sequence. It then disrupts the order of amino acids only within this selected subsequence.

\vspace{0.3em}

\noindent \textbf{Global Reverse} \cite{sun2024enhancing}. It flips the order of all amino acids in the sequence to create a new variant while maintaining the original amino acid composition.

\vspace{0.3em}

\noindent \textbf{Insertion} \cite{sun2024enhancing}. It randomly inserts some amino acids into the protein sequence to create variation.

\vspace{0.3em}

\noindent \textbf{Expansion} \cite{sun2024enhancing}. Inspired by AptaTrans \cite{shin2023aptatrans}, this sequence-level method first finds Frequent Consecutive Subsequences (FCS) in the protein. It then expands the sequence by adding copies of the FCS at a specific location.

\vspace{0.3em}

\noindent \textbf{Repeat Contraction} \cite{sun2024enhancing}. Also inspired by AptaTrans \cite{shin2023aptatrans}, this sequence-level augmentation locates FCS in the protein. It directly deletes some copies of the FCS to shorten the sequence while preserving key patterns.

\vspace{0.3em}

\noindent \textbf{Substitution} \cite{sun2024enhancing}. It adopts a random replacement strategy (since 20 amino acids lack word-like proximity) to substitute some amino acids in the sequence. Here we made some improvements on this operation. First, we categorizes 20 amino acids into six functional groups: hydrophobic/aliphatic (A, V, L, I, M), aromatic (F, Y, W), polar uncharged (S, T, N, Q), negatively charged (D, E), positively charged (K, R, H), and special (G, C, P). When substituting a target amino acid, there are two probabilistic choices: either replace it with another amino acid from the same functional group (to retain chemical properties) or replace it with alanine (A, the simplest aliphatic neutral amino acids. It primarily functions as a ``spacer amino acid'' in protein structures, maintaining the flexibility of the polypeptide chain rather than forming specific functional structures.). The dual-option design balances structural conservation and controlled variation.

\vspace{0.3em}

\noindent \textbf{Mask} \cite{rao2021msa,zhang2023systematic,zhangprotein}. It randomly selects some amino acids in the sequence and replaces them with a special Mask token.

\subsection{3D Structure Operations}

\noindent \textbf{Torsion Angle Perturbation} \cite{lan2025contrastive}. It is a 3D structural augmentation method for ligands, designed to simulate the conformational flexibility of ligands under different solvent conditions. It applies Gaussian noise to the torsion angles of the ligand’s initial 3D structure to generate unbiased, unbound-like torsional topologies, while keeping the ligand’s atomic types unchanged.

\vspace{0.3em}

\noindent \textbf{Gaussian Coordinate Noise} \cite{hu2024deep,hu2024learning,lan2025contrastive}. It is a 3D geometric augmentation method for protein amino acid coordinates, designed to enhance the model’s robustness to small spatial variations of proteins. During the data preprocessing stage, after normalizing the amino acid coordinates, it adds isotropic Gaussian noise to the 3D coordinates of each amino acid, simulating minor fluctuations in protein spatial structure caused by environmental factors.

\vspace{0.3em}

\noindent \textbf{Rotation} \cite{nguyen2025advances}. It rotates the protein's 3D structure by a random angle around a randomly selected axis in 3D space.

\vspace{0.3em}

\noindent \textbf{Translation} \cite{nguyen2025advances}. It shifts the entire set of 3D coordinates of a protein by a small random vector in the spatial domain.

\subsection{1D \& 3D Operations}

\noindent \textbf{Substructure Sampling} \cite{zhangprotein,hermosilla2022contrastive,kalifa2025fusionprot,zhang2023systematic}. It is a method of sampling local sub - structures from protein sequences or 3D structures. For example, it may randomly select a continuous fragment from the protein sequence or a local region from the 3D structure, so that the model can learn more local features and interactions, enhancing the model's recognition ability for specific sub - structures of proteins.

\vspace{0.3em}

\noindent \textbf{Subspace Sampling} \cite{zhangprotein,kalifa2025fusionprot}. It is designed to capture spatially correlated structural motifs (e.g., functional domains with close spatial proximity). It randomly selects a residue $p$ as the center, then extracts all residues within a predefined Euclidean distance radius $d$ from $p$.

\begin{table*}[!t]
  \centering
  \caption{Hyperparameters and their settings for the augmentation operations employed in our experiments.}
  \vspace{-0.5em}
  \scalebox{0.9}{
    \begin{tabular}{l|c|c}
    \toprule
    \multicolumn{1}{c|}{Operation} & Hyperparameter & Range \\
    \midrule
    Substitution & Substitution Ratio & Randomly sampling from $[0.05,0.2]$ \\
    Mask  & Mask Ratio & Randomly sampling from $[0.05,0.2]$ \\
    Torsion Angle Perturbation & Perturbation Angle Range & Randomly sampling from $[-5,5]$ \\
    Gaussian Coordinate Noise & Noise Range & The mean is 0, and the standard deviation range is $[0.05,0.1]$\text{\AA} \\
    Substructure Sampling & Sampling Ratio & Randomly sampling from $[0.6, 0.8]$ \\
    Subspace Sampling & Sampling Radius Range & Randomly sampling $[10.0-20.0]$\text{\AA} \\
    \bottomrule
    \end{tabular}}%
  \label{tab:hyperparameters_operations}%
\end{table*}

\begin{figure*}[!t]
  \centering
  \includegraphics[scale=0.45]{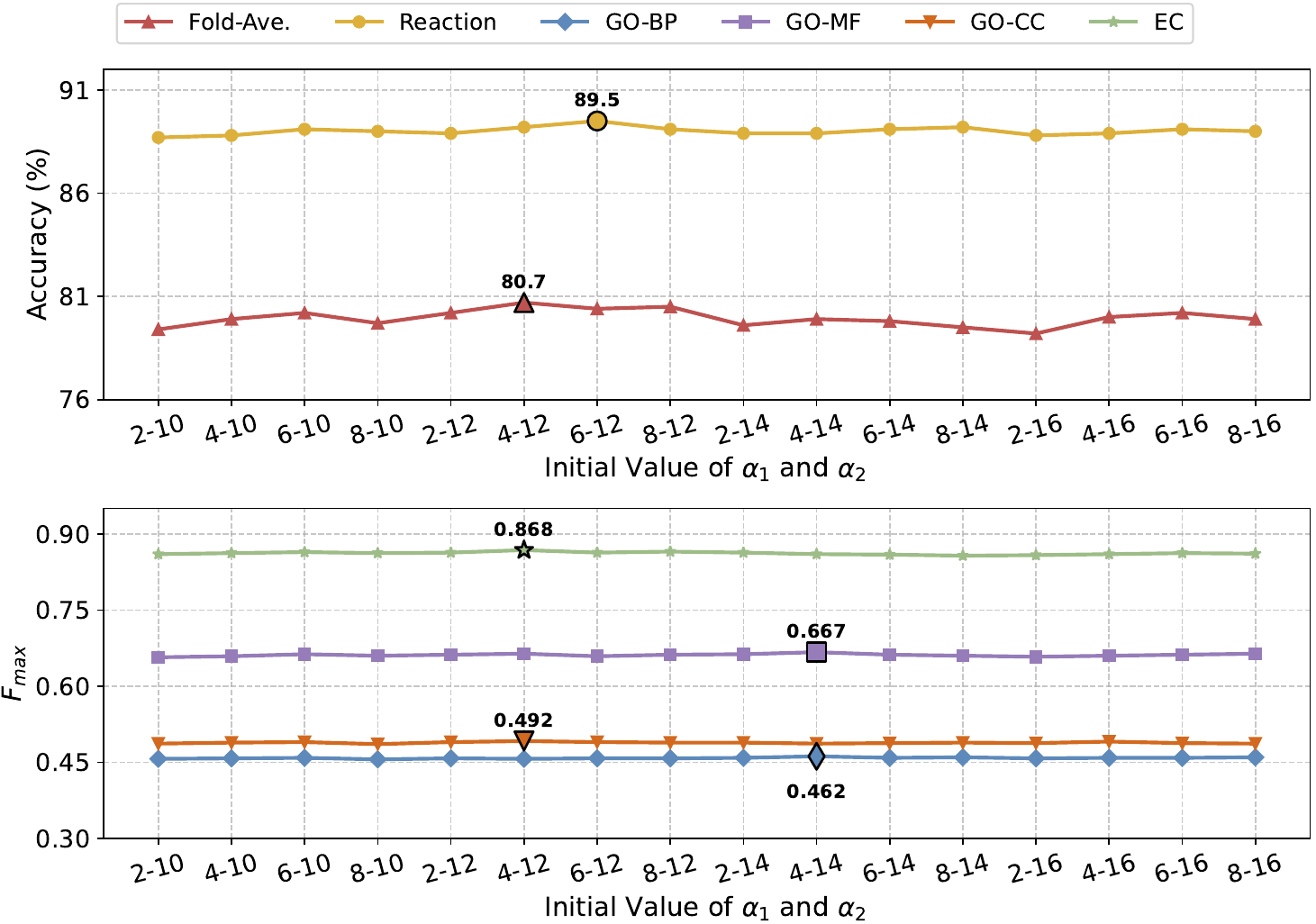}
  \vspace{-0.5em}
  \caption{Effect of initial value $\alpha_1$ and $\alpha_2$ on model performance. For example, ``$2-10$'' represents initial $\alpha_1=2$ and $\alpha_2=10$. We highlight the best performance of each task for ease of reading.}
  \label{fig:alpha}
  \vspace{-0.5em}
\end{figure*}

\section{Experimental Details}
\label{sec:experimental_details}

\subsection{Evaluation Tasks}
\label{sec:tasks}

We evaluated the model's effectiveness across four tasks, adhering to the task and evaluation settings widely adopted in prior work \cite{fan2023continuous,zhangprotein,quan2024clustering}. Specific details are as follows:

\vspace{0.3em}

\noindent \textbf{Protein Fold Classification.} It is a key task in PRL that aims to categorize proteins based on their three-dimensional structural folds, which are crucial for understanding evolutionary relationships and functional similarities. The fold classification task is conducted using the SCOPe 1.75 dataset \cite{hou2018deepsf}, containing 16,712 proteins across 1,195 fold classes. The 3D coordinates of proteins are derived from the SCOPe 1.75 database \cite{murzin1995scop}. The dataset provides three evaluation scenarios: Fold: Excludes proteins from the same superfamily during training; Superfamily: Excludes proteins from the same family during training; Family: Includes proteins from the same family during training. Mean accuracy is used as the metric \cite{wangenhancing}.

\vspace{0.3em}

\noindent \textbf{Enzyme Reaction Classification.} This task predicts the type of chemical reaction catalyzed by an enzyme, based on the four-level Enzyme Commission (EC) hierarchy. The dataset includes 384 four-level Enzyme Commission classes and comprises
29,215/2,562/5,651 proteins for training/validation/test, respectively \cite{fan2023continuous,quan2024clustering}. Mean accuracy is used as the evaluation metric \cite{wangenhancing}.

\vspace{0.3em}

\noindent \textbf{Gene Ontology Term Prediction.} It is a multi-label classification task that assigns proteins to terms in the Gene Ontology (GO) database, which describes functions across three domains: biological process (BP), molecular function (MF), and cellular component (CC). The dataset is organized into three hierarchical ontologies: biological process (BP, 1,943 classes), molecular function (MF, 489 classes), and cellular component (CC, 320 classes). The train/validate/test sets contain 29,898/3,322/3,415 proteins, respectively \cite{fan2023continuous,quan2024clustering}. The evaluation metric is $F_{max}$. 

\vspace{0.3em}

\noindent \textbf{Enzyme Commission Number Prediction.} It is a multi-label task focused on assigning detailed EC numbers to enzymes, covering three- and four-level classifications across hundreds of classes. The train/validate/test sets contain 15,550/1,729/1,919 proteins, respectively \cite{fan2023continuous,quan2024clustering}. Following previous work, for GO term and EC number prediction, we employ the multi-cutoff splits to ensure the test set only includes PDB chains with a sequence identity $\leq$ 95\% to the training set. The evaluation metric is $F_{max}$.

\subsection{Implementation Details}
\label{sec:implementation_details}
We adopt the codes provided by the authors for all baselines. We carefully tune all the hyperparameters as reported in the papers to ensure fair comparison. For our MRM, we tune the initial $\alpha_1, \alpha_2$ and $\gamma$ in the range of $\{2,4,6,8\}$, $\{10,12,14,16\}$, $\{0.2,0.6,1.0,1.4,1.8\}$, respectively. We do not set an additional step size $s$, but instead adopt the simplified implementation of the difficulty scheduler proposed in Section \ref{sec:difficulty}. For hyperparameters in data augmentation operations, we also set them within the ranges recommended in the original paper. We summarize the settings of these operations' hyperparameters in Table \ref{tab:hyperparameters_operations}.

\section{Additional Experimental Results}
\label{sec:additional_results}

\subsection{Hyperparameter Investigation}
\label{sec:more_hyperparameter}

We investigate the impact of initial values $\alpha_1$ and $\alpha_2$ on model performance. The results are illustrated in Figure \ref{fig:alpha}. We can observe that optimal performance typically occurs at ``$4-10$'' and ``$4-14$'' and their vicinity. Both smaller $\alpha_1$ and larger $\alpha_2$ (or an excessively difference between $\alpha_1$ and $\alpha_2$) lead to performance degradation. Appropriate values of $\alpha_1$ and $\alpha_2$ ensure the model is provided with samples that contain sufficient original information and rich variation. As shown in Figure \ref{fig:framework} and Equation (\ref{eq:mix_representation}), an excessively small $\alpha_1$ results in a low proportion of augmented protein data in the final mixed representations, failing to introduce sufficient variation to enhance model performance and robustness. Excessively large $\alpha_1$ reduces the proportion of original protein data in the final mixed representations, causing loss of crucial protein information. Since optimal settings cluster closely together, our method minimizes the need for extensive hyperparameter tuning. In practice, we recommend that users prioritize tuning the initial values of $\alpha_1$ and $\alpha_2$ near the optimal settings shown in Figure \ref{fig:alpha}.


\end{document}